\documentclass[letterpaper, 10 pt, conference]{ieeeconf}  

\IEEEoverridecommandlockouts                              

\usepackage{hyperref}
\hypersetup{
    colorlinks=true,
    linkcolor=blue,
    filecolor=magenta,      
    urlcolor=blue,
}
\usepackage{graphicx,url}
\usepackage{amsmath}
\usepackage{algorithm}
\usepackage[noend]{algpseudocode}
\usepackage{balance}
\usepackage{gensymb}
\usepackage{multirow}
\usepackage{hhline}
\usepackage{array}
\usepackage[font=small,labelfont=bf]{caption}
\usepackage{dblfloatfix}
\usepackage{gensymb}
\title{\LARGE \bf
Improved Object Pose Estimation via Deep Pre-touch Sensing
}

\author{Patrick Lancaster $^{1}$ Boling Yang $^{2}$ and Joshua R. Smith $^{3}$
\thanks{This work was supported in part by the National Science
Foundation under grant IIS-1427419}
\thanks{$^{1}$Patrick Lancaster is with the Paul G. Allen School of Computer Science and Engineering, University of Washington,
        Seattle, WA 98195, USA
        {\tt\small planc509@cs.washington.edu}}%
\thanks{$^{2}$Boling Yang is with the Department of Electrical Engineering, University of Washington,
        Seattle, WA 98195, USA
        {\tt\small bolingy@washington.edu}}%
\thanks{$^{3}$ Joshua R. Smith is with the Paul G. Allen School of Computer Science and Engineering and the Department of Electrical Engineering, University of Washington,
        Seattle, WA 98195, USA
        {\tt\small jrs@cs.washington.edu}}%
}

\begin{document}

\maketitle
\thispagestyle{empty}
\pagestyle{empty}

\begin{abstract}

For certain manipulation tasks, object pose estimation from head-mounted cameras may not be sufficiently accurate. This is at least in part due to our inability to perfectly calibrate the coordinate frames of today's high degree of freedom robot arms that link the head to the end-effectors. We present a novel framework combining pre-touch sensing and deep learning to more accurately estimate pose in an efficient manner. The use of pre-touch sensing allows our method to localize the object directly with respect to the robot's end effector, thereby avoiding error caused by miscalibration of the arms. Instead of requiring the robot to scan the entire object with its pre-touch sensor, we use a deep neural network to detect object regions that contain distinctive geometric features. By focusing pre-touch sensing on these regions, the robot can more efficiently gather the information necessary to adjust its original pose estimate. Our region detection network was trained using a new dataset containing objects of widely varying geometries and has been labeled in a scalable fashion that is free from human bias. This dataset is applicable to any task that involves a pre-touch sensor gathering geometric information, and has been made publicly available. We evaluate our framework by having the robot re-estimate the pose of a number of objects of varying geometries. Compared to two simpler region proposal methods, we find that our deep neural network performs significantly better. In addition, we find that after a sequence of scans, objects can typically be localized to within 0.5 cm of their true position. We also observe that the original pose estimate can often be significantly improved after collecting a single quick scan.

\end{abstract}

\begin{figure}[t]
\centering
\includegraphics[width=0.45\textwidth]{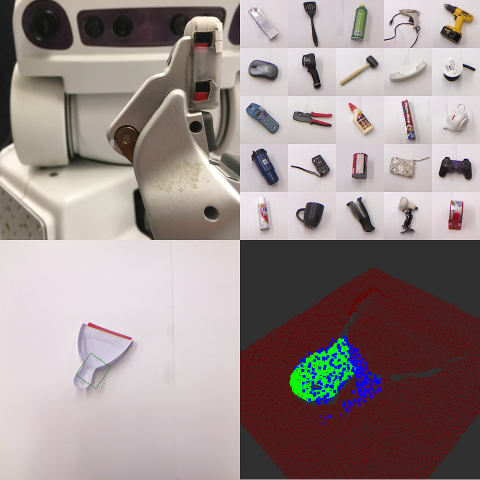}
\caption{ By combining pre-touch sensing with a deep region detection network, the robot can efficiently obtain an improved object pose estimate. Top Left: The robot and the finger-mounted optical time-of-flight pre-touch sensor. Top Right: A subset of the items in the created dataset. Bottom Left: An example annotation from the dataset. Bottom Right: A visualization of the alignment between Kinect data (green) and pre-touch data (blue).
}
\label{fig:front}
\end{figure}
\section{INTRODUCTION}

In order to achieve object manipulation in unstructured environments, a robot must be able to estimate object pose with respect to its end-effectors at an accuracy that is determined by the task. Head-mounted sensors that are commonly used in robotics may not be able to provide sufficient accuracy because of our inability to perfectly calibrate the robot's joints and cameras. Even slight  mis-calibrations can be compounded by the high degree-of-freedom nature of modern robot arms, resulting in a significant positional and/or rotational error. Alternatively, proximity sensors mounted to the robot's end-effectors can avoid these calibration errors, and thereby provide the feedback that is necessary for highly precise tasks. We refer to this type of sensing as 'pre-touch' because it allows the robot to measure the object at close range but before making contact. However, these types of sensors typically only provide spatially sparse measurements, and therefore require more actuation to collect enough data to re-estimate an object's pose. In order to minimize actuation time, we propose to train a deep neural network to predict object regions that will yield discriminative information about the object's pose when scanned by a pre-touch sensor.

\begin{figure*}
  \includegraphics[width=\textwidth,height=5cm]{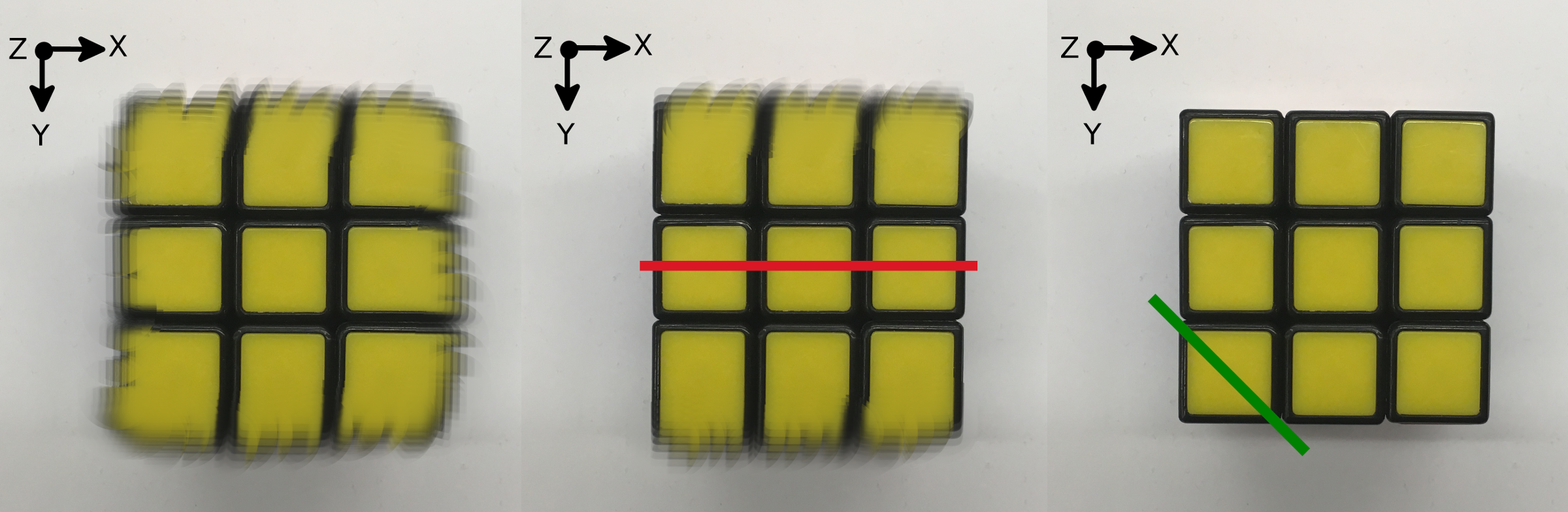}
  \caption{Left: Suppose the pose in the x and y directions is initially uncertain. Middle: A horizontal pre-touch scan localizes the object in the x direction, but its position in the y direction is still uncertain. Right: Alternatively, a diagonal scan at the corner yields additional geometric features that eliminate uncertainty in both the x and y directions.}
  \label{fig:uncertainty}
\end{figure*}

While specifying which regions are optimal for pre-touch sensing is not, in general, obvious, deep neural networks have the potential to learn feature hierarchies that encode them. An example of what such a deep network might learn is illustrated in Fig. \ref{fig:uncertainty}. Suppose that the robot is uncertain about the pose of the object in the x and y directions, but is equipped with a pre-touch sensor that can measure depth along the z axis. Applying the pre-touch sensor across the horizontal region would provide much information about the pose in the x direction, but little information about the pose in the y direction. A more efficient strategy would be to scan diagonally at the corner of the object, allowing the robot to estimate the pose of the object in both the x and y directions well. While this is an example for a simple object, in general we encourage our network to predict regions that are relatively unique with respect to adjacent regions. We expect that such regions will be irregularly shaped, asymmetric and/or contain edges.

The application of machine learning to pre-touch sensing has not been well explored. To the best of the authors' knowledge, this is the first work to use deep neural networks to predict regions that are amenable to pre-touch sensing. We also created a publicly available dataset (\href{https://bitbucket.org/planc509/pretouch_learn_dataset}{https://bitbucket.org/planc509/pretouch\_learn\_dataset}) of 136 common items with widely varying geometries. A subset of these items is shown in Fig. \ref{fig:front}. This dataset is applicable to any pre-touch sensor that gathers geometric information. We are unaware of any other databases for pre-touch sensing of comparable size and variety.

The remaining sections proceed as follows. Section 2 examines related work. Section 3 describes a framework for how pre-touch sensing can be used for better pose estimation. Then section 4 describes the data collection and learning methods that the efficiency of this framework is dependent upon. Section 5 evaluates the overall system and section 6 presents conclusions from this work.

\section{Related Work}

In order to effectively interact with an object in the presence of occlusions and uncertainty, a robot needs to decide how to employ its sensors such that it collects the most informative feedback relative to the current task. Here we examine two proposed components of this process that up to this point have mostly been kept separate. We first review work related to sensors that break the convention of being mounted on the robot's head in an attempt to sense at a closer range and avoid occlusion. In the second part, we explore ways in which deep learning has been applied to robot manipulation.

\subsection{Occlusion Robust Sensing} 
\subsubsection{Eye-In-Hand Sensing} 
One way to reduce occlusion is to simply move conventional head-mounted sensors closer to the robot's end effector. Leeper et al. \cite{leeper2014using} attach a stereo vision system to the robot's wrist in order to improve grasping. The depth map generated by the wrist mounted sensor is used to adjust the final grasp, as well as determine the likelihood of a grasp executed from the current pre-grasp position to succeed. Kahn et al. \cite{kahn2015active} use a wrist mounted RGB-D camera to search for occluded grasp handles. The use of trajectory optimization allows for more effective exploration of the robot's workspace.

\subsubsection{Tactile Sensors}
While wrist and head mounted sensors operate from relatively far away, tactile sensors measure upon making contact with an object.
Petrovskaya and Khatib \cite{petrovskaya2011global} integrate this sensing modality into a bayesian framework for estimating object pose, allowing the robot to robustly grasp objects and open a door. They develop the particle-based Scaling Series algorithm that iteratively estimates the robot's belief of the object's pose given a series of tactile measurements and object mesh model. Schmitz et al. \cite{schmitz2014tactile} train a deep neural network to recognize objects based on tactile feedback. Using the 241 element sensor array that covers their four-fingered robot hand, they demonstrate the vast recognition improvement provided by a deep network compared to a shallow model when evaluated on their twenty item dataset. These works are examples of effective robot tactile sensing, but they all avoid the issue of contact causing incidental displacement of the object and thereby invalidating the robot's belief or even causing the task to fail.

\subsubsection{Pre-touch Sensors}
Pre-touch sensors serve as a compromise between short-range tactile sensors and long-range head mounted sensors. Like tactile sensors, they are attached to the end effector, allowing the robot to deal with occlusion, as well as potentially make more accurate measurements by sensing at a closer range. However, like head mounted sensors, they do not require contact before making a measurement, and thereby overcome the issue of incidental object displacement that tactile sensing can cause. Numerous sensing modalities for pre-touch have been explored, each of which has its own unique benefits and drawbacks.

Optical pre-touch sensors operate by estimating distance through the measurement of infrared beams that are emitted by the sensor and reflected by the object. While the optical modality is applicable to a wide variety of objects, most implementations are not able to provide accurate distance estimations of transparent objects. Hsiao et al. \cite{hsiao2009reactive} estimate the pose of the surface local to each of their robot's fingers using a bayesian belief framework based on optical pre-touch measurements. While \cite{hsiao2009reactive} is limited to local surface estimation, our method is capable of both local and global pose estimation. Maldonando et al. \cite{maldonado2012improving} use an optical sensor to measure areas that are occluded from the head-mounted sensor's view, allowing them to compute better quality grasps. Guo et al. \cite{guo2015transmissive} use a break beam optical sensor to sense objects that are difficult to measure with other optical sensors. Yang et al. \cite{yang2017pretouch} use an optical time-of-flight pre-touch sensor to accurately manipulate a Rubik's cube, as well as roughly estimate the pose of a number of objects.

Electric field sensing has also been widely explored for pre-touch sensors. These sensors typically operate by generating a displacement current from a transmit electrode to one or more receive electrodes. Objects in proximity to the sensor can alter the signal measured at the receiver(s) by either increasing conductance between the electrodes or by shunting current away. Electric field sensors are most effective when measuring objects that have a dielectric constant significantly different than air, such as metal, while items with low dielectric contrast, such as plastic, are difficult to detect. The use of electric field sensing for robot grasping is explored in \cite{smith2007electric} and \cite{wistort2008electric}. This modality has also been applied to other robot tasks, such as estimating the fill level of a container \cite{muhlbacher2015responsive} and collision avoidance \cite{smith2007electric} \cite{schlegl2013virtual}.

Acoustic pre-touch sensing is an alternative to both optical and electric field methods. Jiang and Smith \cite{jiang2012seashell} develop the first acoustic pre-touch sensor, which operates by measuring the resonant frequency of a pipe embedded inside of the sensor. The resonant frequency varies as objects near the pipe cause the effective length of the pipe to change. In \cite{huang2015sensor}, this sensor is combined with haptic-aided teleoperation in order to grasp transparent objects, which are difficult to detect with more common optical-based sensors. However, their acoustic pre-touch sensor is relatively limited in range, and does not detect open foams, rough fabrics, and fur well.

\subsection{Deep Learning for Manipulation}
Deep learning allows robots to learn a feature hierarchy that transforms raw, high-dimensional sensory inputs (such as an RGB-D image) into a lower-dimensional space that represents the task well and in which a decision can more easily be made. Due to the importance of grasping in robotics, the application of deep learning to this particular manipulation task has received much attention. In one of the earliest works in this area, Lenz et al. \cite{lenz2015deep} use two cascaded neural networks to detect grasp points. Redmon and Angelova \cite{redmon2015real} employ convolutional neural networks to improve both the accuracy and run-time of grasp detection. Both \cite{levine2016learning} and \cite{pinto2016supersizing} approach the task of learning how to grasp from data that is not explicitly labeled by a human. In this sense, we adopt a similar approach to collecting data for our task of improving pose estimation.

Deep learning has also been applied to more complex robot tasks. Yang et al. \cite{yang2015robot} learn from internet videos in order to synthesize plans for robotic cooking. By combining two convolutional networks for object recognition and grasp prediction, a high level manipulation module generates appropriate action sequences. Finn et al. \cite{finn2016deep} use reinforcement learning to train controllers that are informed by task features from deep spatial encoders. Using this strategy, they build controllers for pushing objects, scooping with a spatula, and placing a loop onto a hook.

\begin{figure}[t]
\centering
\includegraphics[width=0.45\textwidth]{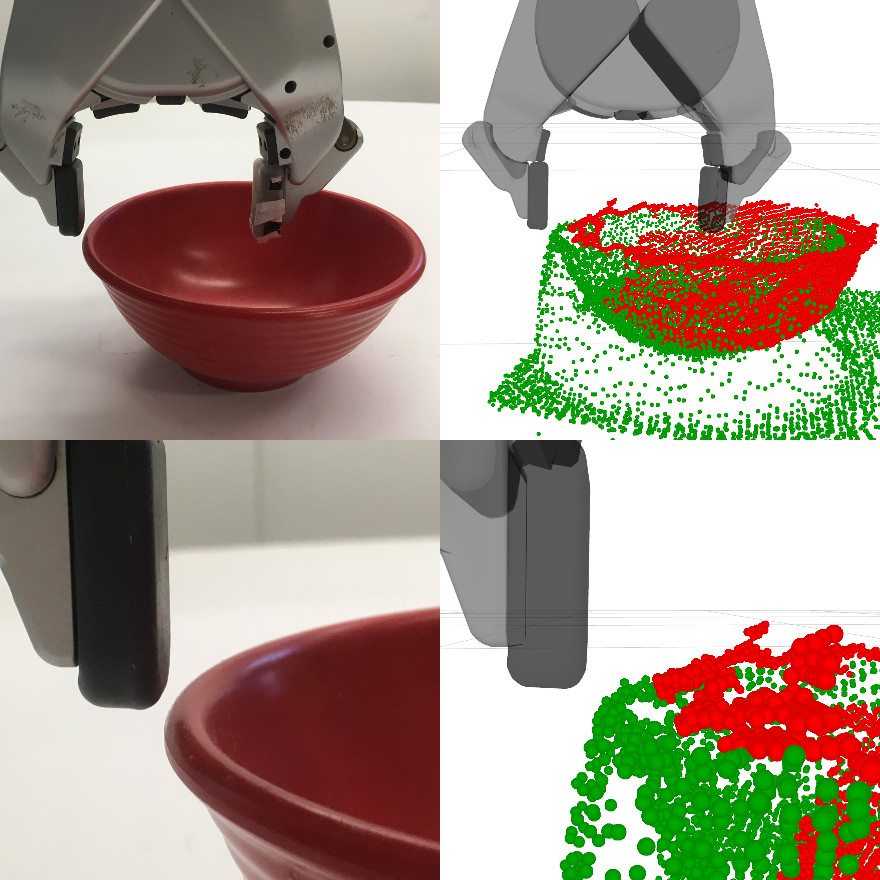}
\caption{Left: A far and close view of the surface of the robot's fingertip aligned with the edge of the bowl. Right: Pre-touch measurements (green) and Kinect measurements (red) with respect to the fingertip. 
}
\label{fig:motivation}
\end{figure}

\section{Pre-touch Sensing for Object Localization}
For some robot manipulation tasks, accurate object pose estimation can be vital to task success. However, head-mounted sensors may not be able to provide sufficient accuracy due to imperfect calibration between the sensor and the robot's end effector. Here, we use an optical pre-touch sensor mounted to the robot's end effector to collect geometric information about the object, and thereby avoid the previous calibration error. We then apply the iterative closest point (ICP) algorithm to obtain a more accurate pose.
\subsection{Motivation}
The use of pre-touch sensors is encouraged by the inability to completely eliminate calibration error between the robot's head mounted sensors and its end effectors. Note that just prior to beginning this work, our robot was calibrated using standard procedures \cite{pradeep2014calibrating}. Fig. \ref{fig:motivation} demonstrates the typical severity of mis-calibration, as well as how pre-touch sensing can provide a better estimate of the pose of the object. While in reality the inner surface of the robot's fingertip is aligned with the edge of the bowl, data from the head-mounted Kinect indicates that they are not. In contrast, the data from the optical pre-touch sensor is significantly closer to reality.  

\begin{figure}[t]
\centering
\includegraphics[width=0.4\textwidth]{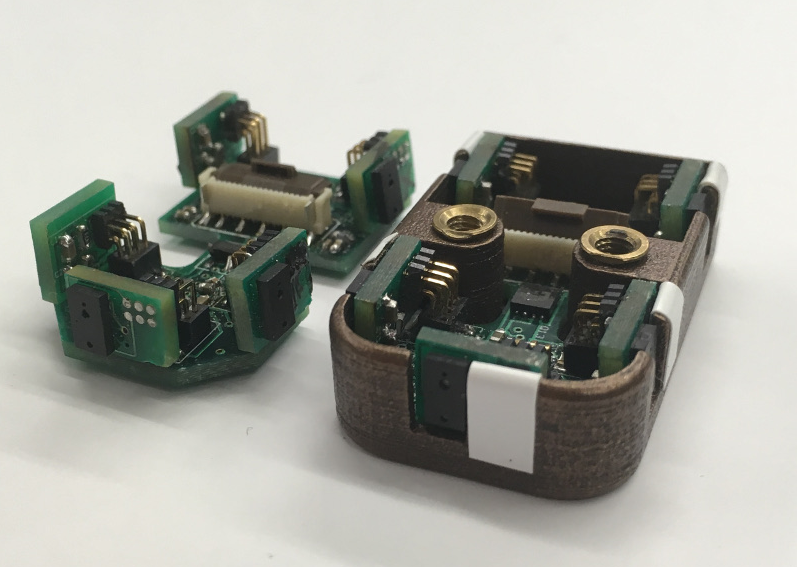}
\caption{The optical time-of-flight sensor, with and without its casing.
}
\label{fig:pretouch_sensor}
\end{figure}

\subsection{Sensor Hardware}
Optical pre-touch sensing can provide accurate distance measurements when applied to a wide range of objects. This work used the optical time-of-flight pre-touch sensor from \cite{yang2017pretouch} and is shown in Fig. \ref{fig:pretouch_sensor}. It is able to measure objects at a range of 1 to 10 cm with millimeter-scale accuracy.  The sensor simultaneously measures from its six sensing modules, each at a rate of 30 Hz. However, only the ranging module at the tip of the finger is used in this work.  

\subsection{Data Registration}
\label{sec:data_registration}
Given a point cloud of an object from a head-mounted Kinect and a point cloud resulting from a scan by the (more accurate) pre-touch sensor, correspondences between the two sets of data must be found in order to quantify the difference between the two. It is expected that the discrepancy between their centroids be on the order of a few centimeters, and that the two clouds have similar overall orientations. Aligning two point clouds that have a misalignment of this magnitude is well suited to the ICP algorithm, as opposed to a full registration approach that might be taken when the pair are more drastically misaligned. In aligning the two clouds using ICP, the algorithm yields a spatial transformation between the robot's old error prone belief (based on the Kinect cloud) and its new belief (based on the pre-touch sensor cloud).   

While the above strategy allows the robot to re-estimate the pose of an object, obtaining a full pre-touch scan of the object can be expensive in terms of actuation time. It would be more efficient if the robot could re-estimate pose by only scanning a portion of the object and then matching it with the corresponding portion of the original Kinect cloud. The difficulty is that as less pre-touch data is gathered, the number of discriminative features in the scan decreases, making the matching more susceptible to error caused by noise and differing sensor characteristics. However, by choosing regions to scan that are rich in discriminative geometric features, the robot can collect data efficiently while still achieving high quality pose re-estimation. We describe how we learn to detect such regions in the next section. 
\\
\\
\\
In this work, we use the Point Cloud Library's (PCL) implementation of ICP \cite{rusu20113d}\cite{holz2015registration}. Beyond PCL's default implementation of ICP, we also use a correspondence rejector that limits the influence of outliers \cite{phillips2007outlier}.

\section{Learning Where to Sense}
We propose to use a deep neural network to detect object regions that are amenable to pre-touch sensing. Such regions will contain discriminative geometric features that allow the robot to accurately re-estimate the pose of the object. We parameterize these regions as rectangles in the image plane, and re-purpose a state-of-the-art object detection neural network for our task. In order to generate examples from which the network can learn, we form a large set of random candidate regions, simulate pre-touch measurement in those regions, and retain the regions that are most robust to random noise and offsets when being matched to the corresponding region of the Kinect cloud.

\subsection{Region Specification}
Rotated rectangles represent a flexible, yet simple way to specify regions. Jiang et al. \cite{jiang2011efficient} first proposed using this representation for robot grasping, and many works have adopted it \cite{lenz2015deep}\cite{pinto2016supersizing}\cite{redmon2015real}. In this work, a specified rectangle represents a region at which to apply pre-touch sensing. Specifically, we perform the pre-touch scan along the perimeter of the region. Each rectangle can be parameterized as follows: 
\begin{center}
$[x,y,w,h,\theta]$
\end{center}

where x and y are the center of the rectangle in pixels, w and h are the width and height of the rectangle, and $\theta$ is the rotation around the center of the rectangle in the image plane. Following \cite{redmon2015real}, in practice we use cosine and sine of twice the angle to represent the rotation due to the symmetry of the rectangle.


\subsection{Neural Network Architecture}
As noted in \cite{redmon2015real}, choosing rectangles to represent areas of interest essentially reduces the problem to object detection (albeit with an additional parameter for rotation), allowing us to take advantage of the powerful tools that have been developed in that area. A common strategy for detection is to classify sub-patches of the image, and then consider the most certain positive classifications to be detections. The naive approach is to individually classify all sub-patches of the image, but this is very inefficient. Lenz et al. \cite{lenz2015deep} address this issue by using a small neural network to evaluate a large pool of candidates before passing the most promising ones to a larger network. We instead use the Faster R-CNN framework \cite{ren2015faster}, which uses a Region Proposal sub-network to indicate to the detection layers which sub-patches are likely to contain a region of interest. While the original Faster R-CNN implementation \cite{girshick2015} was designed for object detection, we expanded it to be able to predict rotated rectangles (this required us to efficiently calculate the overlap between two rotated rectangles, which was achieved with the General Polygon Clipper library \cite{murta2000general}) We initialized the convolutional layers of our network with the 'CNN\_M' pre-trained network from \cite{chatfield2014return}. Given a depth image, our network outputs a list of detected regions, and each region is accompanied by a confidence score.

\subsection{Data Collection}
\label{sec:data_collect}
Generating examples of regions that are amenable to pre-touch sensing is not a trivial task. Ultimately we want to choose regions that when scanned by a pre-touch sensor, produce data that can be reliably matched with the corresponding regions of the original Kinect cloud in the presence of random noise and calibration error. Although a human may be able to label some of these areas, his or her intuition may not always coincide with what is optimal from an ICP perspective. Furthermore, if we can automate the data collection process, it will be easier to scale up the size of the dataset such that there is sufficient data to avoid overfitting. Here, we generate a large number of random candidate regions, and then build a set of labels based on which regions yield the most robust ICP matching results. We evaluate a large number of candidates by simulating pre-touch measurement of each region.

The label generation algorithm is summarized in Algorithm \ref{rec_gen}. A candidate set of of one thousand rectangles ($n_R = 1000$) is first randomly generated, where each rectangle has an intersection-over-union of at least fifteen percent with the rectangle bounding the object, and has an area that is ten to fifty percent of that of the bounding rectangle. In the \textit{for} loop on line 5, the algorithm builds a target cloud for each rectangle by retaining the subset of points from $K$ whose projections into the image plane are near or on the perimeter of the rectangle. The algorithm then generates ten random ($n_{trials}=10$) offsets, where the translations in the x, y, and z directions are drawn from a uniform distribution with limits of $-2.0$ cm to $2.0$ cm, and roll, pitch, and yaw rotations are drawn from a uniform distribution with limits of $-5.0\degree$ to $5.0\degree$. For each of these generated offsets, the \textit{for} loop on line 10 applies it to $K$ in order to generate $K'$. $K'$ represents a simulation of what a pre-touch sensor would measure if it were applied to the whole object. Then the source cloud for each rectangle is built by retaining the subset of points from $K'$ whose projections into the image plane are on the perimeter of the rectangle. This simulates a noiseless pre-touch scan along the perimeter of the region that corresponds to the rectangle. Before matching each source and target pair using ICP, the algorithm adds gaussian noise ($\sigma=0.15$cm) to the source. The pair is scored by calculating the average distance between the predicted aligned cloud's and the ground truth aligned cloud's points (where the ground truth cloud is computed by applying the inverse of the original offset between $K$ and $K'$ to the noised source cloud). The final score for each rectangle (computed in the \textit{for} loop on line 19) is the worst (i.e. largest) score for that rectangle across all trials. Therefore, in order to get a good (i.e. low) score, a rectangle has to perform well across multiple simulations, which adds robustness to the generated labels. The rectangle with the best score is retained as a label.

\begin{algorithm}
\caption{Label Generation}\label{rec_gen}
\begin{algorithmic}[1]
\Procedure{LabelGeneration}{$K$}\Comment{Kinect cloud K}
\State $R \gets n_R \text{ random rectangles}$
\State $K2D \gets \text{projection of K into image plane}$
\State $T \gets n_R \text{ empty target clouds}$
\For{$\texttt{r in 1:}n_R$}
\For{$\texttt{i in 1:}n_{K2D}$}
\If{$R^r.nearPerimeter(K2D_i)$} 
\State $T^r.addPoint(K_i)$
\EndIf
\EndFor
\EndFor
\State $\Delta K \gets n_{trials} \text{ random translations \& rotations}$ 
\For{$\texttt{j in 1:}n_{trials}$}
\State $K' = K + \Delta K^j$ \Comment{Offset cloud $K'$}
\State $K'2D \gets \text{projection of K' into image plane}$
\State $S \gets n_R \text{ empty source clouds}$
\For{$\texttt{r in 1:}n_R$}

\For{$\texttt{i in 1:}n_{K'2D}$}
\If{$R^r.onPerimeter(K'2D_i)$} 
\State $S^r.addPoint(K'_i)$
\EndIf
\EndFor
\State $trialScores^{j,r} = Score(S^r+N(0,\sigma),T^r)$
\EndFor
\EndFor
\For{$\texttt{r in 1:}n_R$}
\State $scores^r = \max_j (trialScores^{j,r})$
\EndFor
\State $\texttt{return }FilterRecs(R, scores)$
\EndProcedure
\Procedure{Score}{$P,Q$}
\State $PAlign = icp(P,Q)$
\State $\texttt{return } \frac{\sum \lvert PAlign_i-PAlign^*_i \rvert_2}{n_P}$
\EndProcedure
\end{algorithmic}
\end{algorithm}


\begin{table*}
\centering
\begin{tabular}{ |c||c|c|c||c|c|c|  }
 \hline
 &\multicolumn{3}{c||}{Mean Error (cm)} & \multicolumn{3}{c|}{Std. Dev. of Error (cm)}\\
\cline{2-7}
 & Random & NARF & Deep Pre-touch&Random&NARF&Deep Pre-touch\\
 \hline
 Air Freshener	&1.31	&2.21			&\textbf{1.26}	&0.52			&1.50			&\textbf{0.50}\\
 Clock			&2.21	&2.97			&\textbf{0.80}	&1.40			&2.16			&\textbf{0.32}\\
 Cleanser		&1.93	&\textbf{0.71}	&1.16			&1.46			&\textbf{0.58}	&0.60\\
 Controller		&2.77	&2.91			&\textbf{1.49}	&1.33			&1.96			&\textbf{0.76}\\
 Fruit Bowl		&1.69	&1.87			&\textbf{0.83}	&1.50			&1.46			&\textbf{0.29}\\
 Gripper		&2.24	&1.48			&\textbf{1.11}	&1.39			&0.95			&\textbf{0.41}\\
 Toy			&2.55	&1.94			&\textbf{1.72}	&1.13			&0.66			&\textbf{0.43}\\
 Wallet			&1.16	&1.70			&\textbf{0.89}	&\textbf{0.63}	&0.83			&0.66\\

 \hline
\end{tabular}
\captionsetup{font=small}
\caption{Mean and standard deviation pose estimate error across ten individual scans for each of the eight different objects and each of the three methods. All values have centimeter units, and the best value across each of the three methods is bolded. }
\label{table:1}
\end{table*}

\begin{figure*}[!b]
  \includegraphics[width=\textwidth,height=5cm]{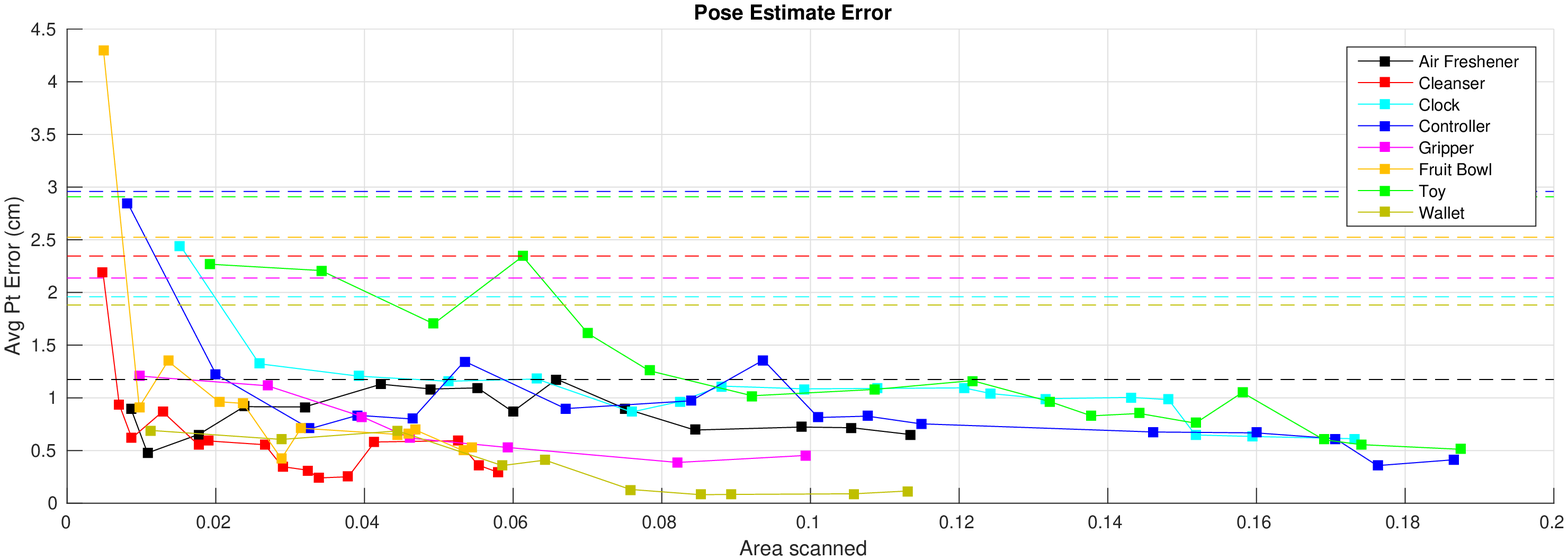}
  \caption{The pose estimate error after each pre-touch scan. The y-axis is the average distance between the ground truth alignment's points and the estimated alignment's points, and the x-axis denotes the percentage of the object that has been scanned. Each object corresponds to a different color, and each square represents a scan. The correspondingly colored dashed lines for each object represent the average distance between the points of the entire ground truth alignment and the points of the whole original Kinect point cloud.}
  \label{fig:icp_fit}
\end{figure*}

\section{Results}

In order to evaluate the proposed framework, we apply it to pose estimation of eight objects that are not in the dataset used to train the deep neural network. For each object, the PR2 robot receives an initial point cloud from the Kinect, which our detection network uses to propose a set of pre-touch scanning regions.  For each region, the robot scans along the path that corresponds to the perimeter of the rectangle. After performing each scan with the pre-touch sensor, we match it with the corresponding region of the Kinect cloud, where the corresponding region is computed in a fashion similar to that of lines 6-8 of Algorithm \ref{rec_gen}. We  obtain a ground-truth estimate of the pose by matching a separate pre-touch scan of the whole object with the Kinect cloud. We argue that this is reasonable because, again, the data from the pre-touch sensor is not affected by mis-calibration of the coordinate frames of the robot's arms. We have also observed, for example as in Fig. \ref{fig:motivation}, that data from the pre-touch sensor matches reality better than data from the Kinect. All scans were performed such that the sensing beam was normal to the supporting table for simplicity, but future systems could use more complex scanning strategies.

\begin{figure*}
  \centerline{\includegraphics[width=16.0cm]{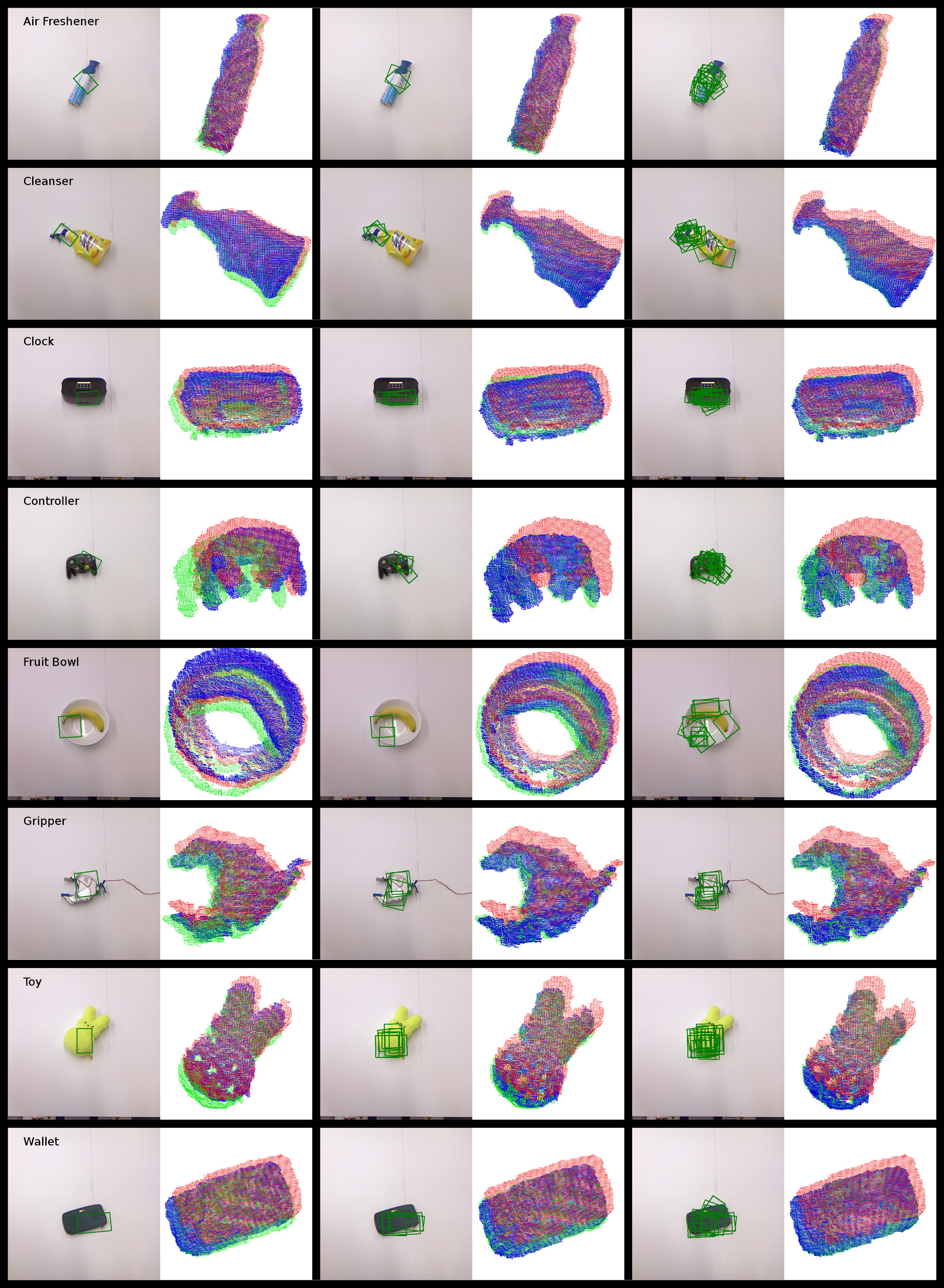}}
  \caption{The matching results throughout the sequence of regions scanned by the pre-touch sensor for each object. For each pair, the left image indicates the regions to be scanned with green rectangles, while the right image displays the result of performing the scans. Three clouds are shown. Red represents the original Kinect data, blue represents the alignment estimated using the scans up to that point, and green represents the ground-truth alignment. The left-most pair corresponds to a single scan, the middle pair corresponds to a few scans, and the right-most pair corresponds to the point at which further scans provided no significant improvement in pose accuracy.}
  \label{fig:icp_viz}
\end{figure*}

\subsection{Individual Scan Comparison}

We compare our region proposal network to two other baseline proposal methods. The first baseline simply proposes random rectangles that have the same properties as those used in the label generation process (Section \ref{sec:data_collect}). The second baseline, which is a variant of normal aligned radial features (NARF)\cite{steder2011point}, proposes regions in which there is a large change in the object's surface. This method first calculates standard NARF keypoints and corresponding descriptors. For each key point, the corresponding rectangle is defined by two perpendicular vectors whose origins are at the keypoint. The direction of the first vector corresponds to the element of the descriptor that has the greatest value, and the direction of the second is chosen according to which of the two elements perpendicular to the first has the greater descriptor value. The length of these vectors is chosen so that the rectangle's area is half that of the rectangle bounding the object. The rectangle's confidence score is the sum of the two elements that correspond to the chosen directions. 

For each of the three methods, we collected scans corresponding to the first ten proposed rectangles for each of the objects, individually matched them to the corresponding portions of the Kinect cloud as previously described, and computed the average distance across corresponding points of the estimated aligned cloud and the ground truth aligned cloud. These pose estimate errors' means and standard deviations are shown in Table \ref{table:1}. The standard deviations can be interpreted as a measure of each method's consistency, where a lower value indicates greater consistency. For all but one of the items, our region detection network had the lowest mean, typically by a significant margin. Also, the network had the lowest standard deviation for most of the items, while being outperformed by a small margin on two of the items.  

\subsection{Deep Pre-touch Sequential Scanning}
While the previous section demonstrated the performance of individual scans proposed by the region detection network, it is possible that a smaller margin of error is desired. We hypothesize that even better pose estimation can be achieved by concatenating multiple scans together. Specifically, we create a sequence of regions to be scanned by ordering them by their confidence scores. After performing each scan with the pre-touch sensor, we concatenate it with the previous scans before matching it with the corresponding regions of the Kinect cloud. 

The results of the pose estimation for each object are shown in Fig. \ref{fig:icp_fit} and Fig. \ref{fig:icp_viz}. As seen in Fig. \ref{fig:icp_fit}, the pose estimation for each object typically improves as additional area is scanned. Fig. \ref{fig:icp_fit} also shows that most of the objects could be localized to within approximately 0.5 cm after scanning less than twenty percent of the object. Some objects, such as the gripper, wallet, and cleanser achieved this accuracy after scanning less than ten percent of the object. Finally, for most of the objects, the pose estimation improves from the Kinect-based pose estimate after only a single scan. The sequence of region detections and pose estimations for each object is shown in Fig. \ref{fig:icp_viz}. Most of the regions proposed by the network seem reasonable. It is especially promising that when examining the cleanser, the first region proposed by the network was the geometrically unique dispersal end of the object. Fig. \ref{fig:icp_viz} further illustrates how a robot could strategically choose how much time it spends scanning. If the robot can only afford a small amount of actuation time, a single scan could be used to most likely improve the pose estimate. If more time is available, the robot can likely achieve a highly accurate pose with just a few scans. On the other hand, if a high amount of accuracy is absolutely required, the robot can do many small scans while avoiding an exhaustive scan of the whole object.

\section{Conclusion} 
\label{sec:conclusion}

We presented a framework for combining pre-touch sensing with deep learning to improve object pose estimation. By using a deep neural network to detect regions that are rich in geometric features, we can obtain enough information to re-estimate the pose without scanning the whole object. In fact, we found that a single scan was often sufficient to improve the estimate. Our region detection network outperformed two baseline methods, and can typically estimate the pose of an object to within 0.5 cm after a sequence of scans. The network is trained using a new dataset that contains minimal human bias and is applicable to any pre-touch sensor that gathers geometric information. This dataset has been made publicly available.

In future work, we would like to improve the accuracy of our method and integrate it into larger manipulation systems. With respect to accuracy, in this work, our network predicted regions that were individually well suited for pre-touch scanning. It is possible that a model that has memory or explicitly considers a sequence of regions could have better performance, such as a recurrent neural network. We also intend to examine the features learned by our model in more detail, as well as compare them to the features learned in other robot tasks, such as object detection and recognition.






\bibliography{bibtex/bib/IEEEabrv.bib,bibtex/bib/IEEEexample.bib}{}
\bibliographystyle{IEEEtran}

\end{document}